%% file: apps_detection.tex
\documentclass[letterpaper, 10 pt, conference]{config/ieeeconf}
\IEEEoverridecommandlockouts
\usepackage{paralist}
\usepackage{float}
\usepackage{booktabs}
\usepackage{tabularx}
\usepackage{soul}
\usepackage{hyperref}
\hypersetup{
    hidelinks,             % Hides all link borders (optional)
    linkcolor=black,       % Keeps section references unlinked
    colorlinks=true,       % Enables colored links
    urlcolor=blue          % Sets URL color
}
\makeatletter
\let\MYcaption\@makecaption
\makeatother
\usepackage[font=footnotesize]{subcaption}
\makeatletter
\let\@makecaption\MYcaption
\makeatother

\usepackage{config/urwchancal}
\input{Mahony_tex_definitions/tex/preamble_compatible}

\input{Mahony_tex_definitions/tex/notation_defs}

\title{\LARGE \bf
Asynchronous Multi-Object Tracking with an Event Camera}

\author{Angus Apps, Ziwei Wang, Vladimir Perejogin, Timothy L.~Molloy and Robert Mahony % <-this % stops a space
\thanks{A. Apps, Z. Wang, T.L.~Molloy and R. Mahony are with the Systems Theory and Robotics Group, Australian National University, (e-mail: \{angus.apps, ziwei.wang1, timothy.molloy, robert.mahony\}@anu.edu.au). V. Perejogin is with the Defence Science and Technology Group (e-mail: vladimir.perejogin@defence.gov.au).}%
\thanks{This research was supported by the Centre for Advanced Defence Research in Robotics and Autonomous Systems (CADR-RAS) project UA216424-S08 ``Variable topology distributed collaborative localisation and control''.}
}

\begin{document}
\baselineskip=0.98 \normalbaselineskip

%%%%%%%%%%%%%%%%%%%%%%%%%%%%%%%%%%%%%%%%%%%%%%%%%%%%%%%%%%%%%%%%%%%%%%%%%%%%%%%

\newcommand{\redline}{\vspace{2mm}\todo{\hrule}\vspace{2mm}}

%%%%%%%%%%%%%%%%%%%%%%%%%%%%%%%%%%%%%%%%%%%%%%%%%%%%%%%%%%%%%%%%%%%%%%%%%%%%%%%
\maketitle
%%%%%%%%%%%%%%%%%%%%%%%%%%%%%%%%%%%%%%%%%%%%%%%%%%%%%%%%%%%%%%%%%%%%%%%%%%%%%%%

\begin{abstract}
    Events cameras are ideal sensors for enabling robots to detect and track objects in highly dynamic environments due to their low latency output, high temporal resolution, and high dynamic range.
    In this paper, we present the Asynchronous Event Multi-Object Tracking (AEMOT) algorithm for detecting and tracking multiple objects by processing individual raw events \emph{asynchronously}.
    AEMOT detects salient event blob features by identifying regions of consistent optical flow using a novel \textit{Field of Active Flow Directions} built from the \emph{Surface of Active Events}.
    Detected features are tracked as candidate objects using the recently proposed Asynchronous Event Blob (AEB) tracker in order to construct small intensity patches of each candidate object.
    A novel learnt validation stage promotes or discards candidate objects based on classification of their intensity patches, with promoted objects having their position, velocity, size, and orientation estimated at their event rate.
    We evaluate AEMOT on a new \textit{Bee Swarm Dataset}, where it tracks dozens of small bees with precision and recall performance exceeding that of alternative event-based detection and tracking algorithms by over 37\%.
    Source code and the labelled event \textit{Bee Swarm Dataset} will be open sourced.
    \footnote{\url{https://github.com/angus-apps/AEMOT}}

    % \todo{[ZW: it is a bit risky to claim the two compared algorithms are state-of-the-art.][TM: agreed, I've toned it down. ]}
\end{abstract}

%-------------------------------------------------------%
%-----                Introduction                 -----%
%-------------------------------------------------------%
\section{Introduction}
\vspace{-0.2mm}
Vision is arguably the most effective sensing modality for autonomous robotic systems operating in most highly dynamic unstructured real-world environments \cite{2011_Corke_robotics}.
Detecting and tracking an unknown number of fast moving, high saliency, visual features
% in scenarios with background activity and clutter
% \cite{2015_vo_multitarget}
is directly linked to the functionality and safety of autonomous systems.
The low latency response (microsecond) and high dynamic range (greater than 120dB) of event cameras make them an ideal sensor to provide this capability \cite{2019_falanga_fast-perception}.
However, event camera data is \emph{asynchronous}, a fundamentally different data modality than provided by classical \emph{frame-based} vision sensors, requiring a new paradigm of processing algorithms
\cite{2020_gallego_event-survey}.

\begin{figure}
    \centering
    \subfloat{\label{subfig:a} \includegraphics[width=0.9\columnwidth]{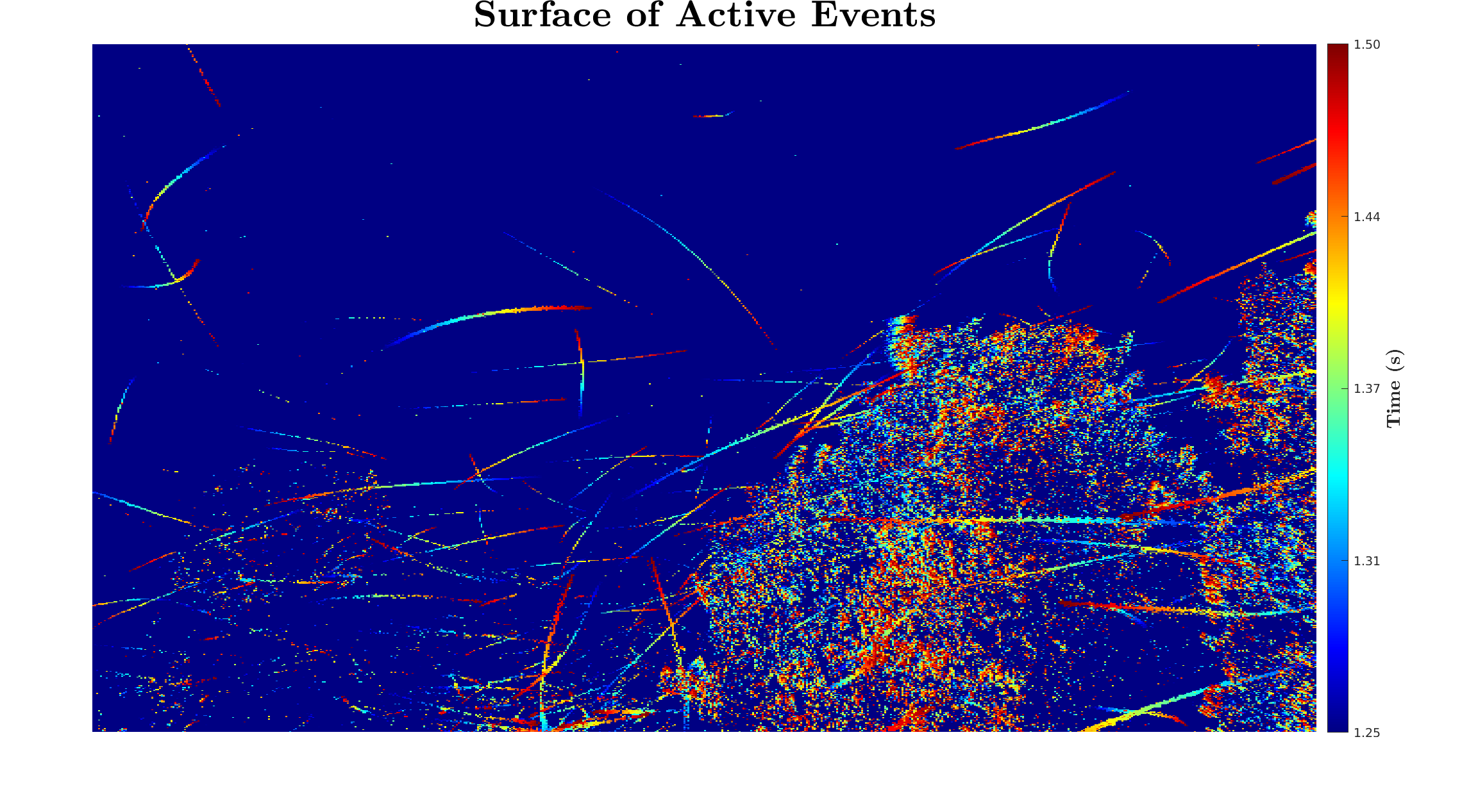}}\\
    \vspace{-4mm}
    \subfloat{\label{subfig:b} \includegraphics[width=0.48\columnwidth]{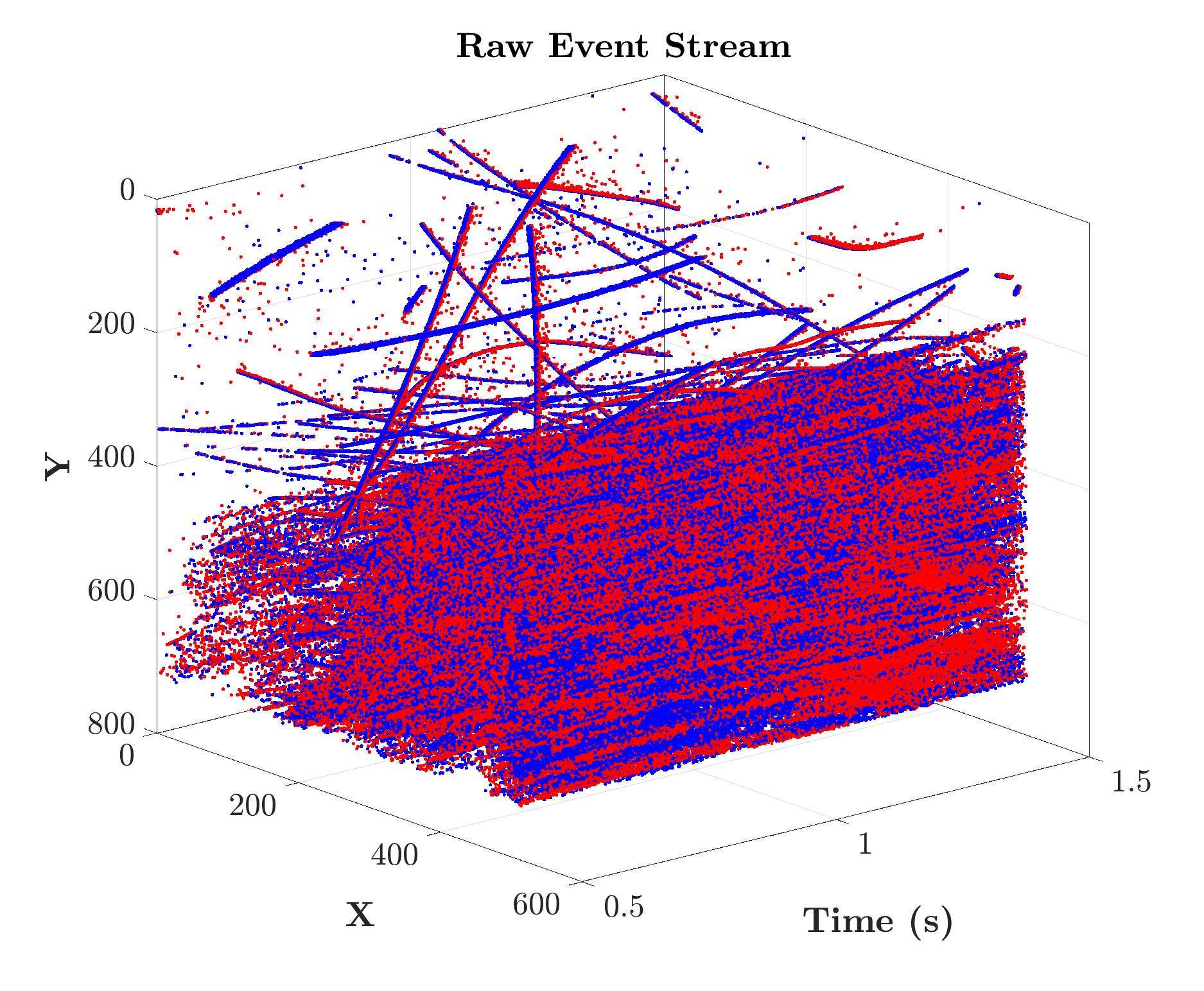}}
    \subfloat{\label{subfig:c} \includegraphics[width=0.48\columnwidth]{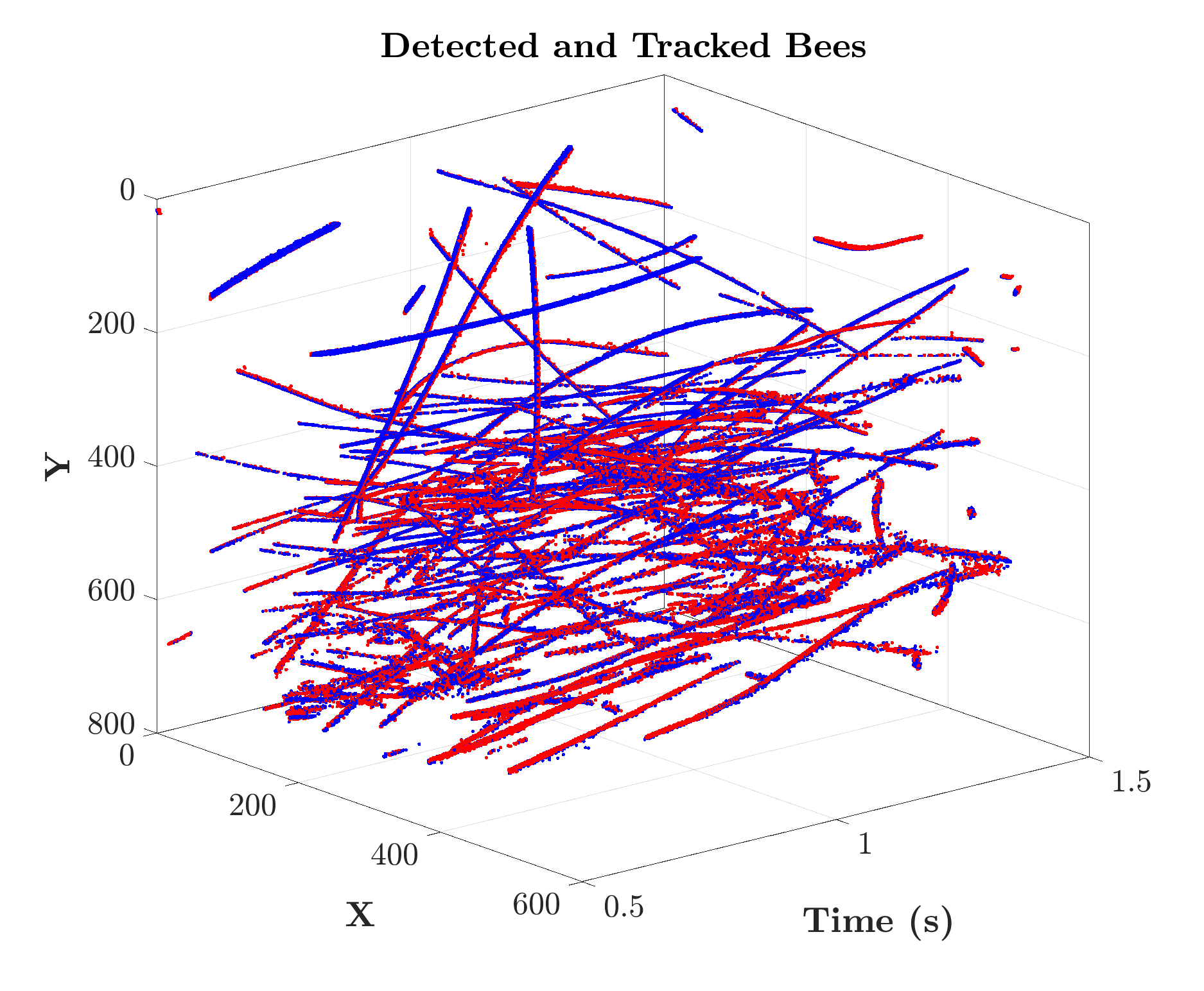}}
    \vspace{-3mm}
    \caption{Our Asynchronous Event Multi-Object Tracking (AEMOT) algorithm effectively detects and tracks dozens of bees in the \emph{Bee Swarm Dateset} using an \emph{asynchronous} detection, validation, and tracking approach. This approach provides robustness against background activity and enables the detection, validation, and tracking of multiple objects at the temporal resolution of their individual event rates}
    \label{fig:aemot_cover}
    \vspace{-6mm}
\end{figure}

Vision-based object tracking is a well established field in robotics with a long history \cite{2021_luo_mot_review}.
Tracking objects using event cameras has received interest recently with applications in particle tracking and velocimetry \cite{2011_drazen_particles}\cite{2020_wang_particle-velocimetry}, obstacle avoidance \cite{2020_falanga_avoidance}\cite{2020_sanket_evdodgenet}, interception \cite{2013_delbruck_robotic-goalie}, and simultaneous localisation and mapping (SLAM) \cite{2016_kim_tracking-and-reconstruction}\cite{2022_mahlknecht_vio}.
Existing tracking algorithms can be classified by whether they use a ``window-based'' or ``asynchronous'' event processing methodology.

Window-based tracking algorithms \cite{2011_drazen_particles}\cite{2022_hu_ecdt} perform operations on a sequence (or window) of events.
This improves contextual information, since direct integration of sequences of events provides image-like information, and allows more classical image processing algorithms to be applied.
However, window-based algorithms increase output latency and compromise response time.
In contrast, asynchronous algorithms \cite{2018_alzugaray_ace}\cite{2020_alzugaray_haste}\cite{wang2023event} process events as they arrive.
This reduces latency as tracks can be updated immediately, however, individual events contain little or no contextual image information making many image processing tasks difficult or impossible.

The task of object detection is a particular challenge for asynchronous algorithms since it depends on information from a (spatial) region in the image and does not naturally lend itself to an asynchronous, event-by-event implementation.
Clustering techniques have been proposed \cite{2010_schraml_clustering}\cite{2018_barranco_clustering},
to find areas of high event activity on the image plane, however, classification of detections as specific features or objects remains challenging.
Feature-specific detectors have been developed, such as lines \cite{2016_brandli_elised}, corners \cite{2016_vasco_fast-Harris}\cite{2017_mueggler_fast-corners}, and circles \cite{2016_glover_hough}.
Deep learning methods have also been explored to detect similar features \cite{2019_manderscheid_learnt-corners} and have been extended to more complex classification tasks \cite{2023_gehrig_recurrent-vision-transformers}\cite{2023_lui_detection}.
Although these methods are more selective and are better suited to rejecting background activity, they are window-based algorithms with the inherent latency limitations and are hand-crafted to specific objects.
High-speed general-purpose multi-object detection and tracking with event cameras requires computationally light-weight algorithms to identify and discern salient features from background activity, and maintain robust tracks in highly-dynamic environments with a large number of potential features and objects.

% \todo{[RM: rerplace this last sentence by one that focuses on the problem the way we want to solve it.]}

%% \todo{this paragraph is about event-based tracking}
%
%In parallel to detection,
%
%
%End-to-end multi-object tracking systems for generic features using event cameras have not been deeply explored, particularly for high-speed applications with tens or hundreds of objects with complex backgrounds.
%In these scenarios, fast and light-weight detection and tracking are paramount.
%This
%
%
%\todo{end this paragraph with commentary about the challenges of tracking multiple objects using event cameras (limited prior work, data association, track management when there is no natural synchronous timing of trackers...)}
%

In this paper, we propose the Asynchronous Event Multi-Object Tracking (AEMOT) algorithm to detect and track multiple fast-moving objects using event camera data, shown in Figure~\ref{fig:aemot_cover}.
The core contribution of the paper is the development of an asynchronous blob detection and object classification algorithm for event camera data that couples to the recently proposed Asynchronous Event Blob (AEB) tracker \cite{wang2023event} with track validation and termination, to provide a systematic process to detect, validate, and track multiple objects using event data.
A secondary contribution is the labelling and release of a new \emph{Bee Swarm Dataset} for evaluating multi-object tracking algorithms for event camera data.

Within AEMOT, each new event received is either used to detect new candidate objects, or allocated to an existing candidate or valid object track based on a stochastic model of event blobs underlying the AEB tracker \cite{wang2023event}.
Asynchronous event blob detection is achieved by computing a novel representation of local optical flow, that we dub the \textit{Field of Active Flow Directions}.
The field of active flow directions is computed by performing regression on the \emph{Surface of Active Events} (SAE)  \cite{2013_benosman_sae}, which is an image of the most recent time-stamps received at each pixel that provides local contextual information.
Both the SAE and field of active flow directions are computationally cheap to maintain.
Statistical correlation of the field of active flow directions with recent flow estimates indicates a moving event blob.
The correlation test threshold is set low to ensure that all salient blobs spawn a local AEB tracker \cite{wang2023event} and have an associated 28$\times$28 intensity patch constructed \cite{2018_Scheerlinck_ComplementaryFilter}\cite{2023_Wang_AKF}.
The intensity patches are input into a simple lightweight binary classification neural network to validate or reject candidate objects.
This validation can be run in parallel to the detection module and AEB trackers: this way, all salient features are tracked asynchronously, and validated after sufficient data is accumulated to make a reliable determination of the nature of objects.
%This also ensures that the more computationally intense \todo{classification Neural Network} is not running event by event.
Based on the validation, the instance of the AEB tracker is deleted if invalid or promoted to a valid track which continues to be tracked.
Track management is performed to manage object track lifetime and interaction.

% The AET tracks the new event blob and provides position and velocity estimates of its motion, as well as, confidence measures in the form of covariance of the velocity.
% These confidence measures are then used to promote detected tracks to identified tracks, that continue to be tracked by the AET.
% \todo{add blah about performance etc. }
% \todo{
% \begin{itemize}
%     \item A complete asynchronous mult-target tracking algorithm using an event camera.
%     \item High-speed blob detector using the surface of active events
%     \item \todo{Bee swarm dataset for event camera multi-target tracking}
% \end{itemize}
% }

This paper is structured as follows.
Section \ref{sec:problemFormulation} defines the event-blob model used to describe general blob-like objects.
Section \ref{sec:detection} introduces detection, Section \ref{sec:pretrack_and_validation} defines the pre-tracking and validation of candidate tracks, and Section \ref{sec:tracking} describes the AEB Tracker and track management procedure.
Section \ref{sec:experimental_setup} introduces our new \emph{Bee Swarm Dataset} and methodology for evaluating event-based multi-object detection and tracking algorithms.
Experimental results are provided in Section \ref{sec:experimental_results}, with conclusions in Section \ref{sec:conclusion}.

%-------------------------------------------------------%
%-----             Problem Formulation             -----%
%-------------------------------------------------------%
\section{Problem Formulation}
\label{sec:problemFormulation}

Consider the visual image of an object moving across the field of view of an event camera.
The event stream received is the sequence of events $\{e_k\} =\{(\xi_{k}, \sigma_{k}, t_{k})\}$ with pixel position $\xi_k \in \R^2$,  polarity $\sigma_k \in \{\pm1\}$, and time-stamp $t_{k}$.
An event stream generates an \emph{event blob} when the probability of each event in the sequence occurring is given by a Gaussian distribution (or more generally a blob-like bell shaped distribution) around a moving center \cite{wang2023event}.
Following AEB tracker \cite{wang2023event}, we parameterise the state of an event blob by $\zeta(t) = (p(t),v(t), \theta(t), q(t), \lambda(t), \Delta(t))$, where $p=(p_{x},p_{y})$ is the mean of the Gaussian distribution, $v = (v_x,v_y)$ is the velocity of $p$, $\theta$ is its orientation with angular velocity $q$, $\lambda = (\lambda_1,\lambda_2)$ is its shape (principal correlations).
% \todo{[ZW: why not $\Delta(t)$? It is time varying and with (t) in Eq (1)]}
The state parameter $\Delta = (\Delta_1,\Delta_2)$, termed the  \emph{polarity offset}, captures information on whether the visual object is light against a dark background or dark against a light background and is explained below.
We assume that the uncertainty in the time-stamp measurement is negligible and we condition on $t = t_k$.
We also assume that there is no uncertainty in the polarity measurement $\sigma = \sigma_k$.
The probability $\textbf{p}(e_{k}|\zeta(t_k))$ of an event $e_k$ occurring conditioned on the state $\zeta(t_k)$ at time $t = t_k$ with polarity $\sigma = \sigma_k$ is given by
\cite{wang2023event}
\begin{align}
\textbf{p}(e_{k}|\zeta(t_k)) &:=  \frac{1}
{2\pi\det(\Lambda(t))} \exp(-\frac{1}{2}\tilde{\xi}_k^{\top} \Lambda(t)^{-2}\tilde{\xi}_k)
\label{eq:blob_probability} \\
\tilde{\xi}_k & = \xi_k-p(t_k)-\sigma_k \Delta(t_k) \notag
\end{align}
where
\begin{align}
    \Lambda(t) := R(\theta(t))
    \begin{pmatrix}
        \lambda^{1}(t) & 0 \\ 0 & \lambda^{2}(t)
    \end{pmatrix}
    R^{\top}(\theta(t)).
\label{eq:blob_Lambda}
\end{align}
Here $\Lambda(t)\in \mathbb{R}^{2\times2}$ represents the shape of the object and acts as the covariance of the Gaussian probability.
Note that the polarity offset parameter $\Delta$ is critical, since visual objects generate leading and trailing edges as they move across a background (Figure~\ref{fig:intensity_patch}).
A bright object moving over a dark background generates positive polarity events at the leading edge $\xi_k-p(t_k)-\Delta(t_k)$ and negative events at the trailing edge $\xi_k-p(t_k)+\Delta(t_k)$.
A dark object on a light background will swap the polarity of events received.
Figure~\ref{fig:intensity_patch} shows the bimodal distribution of positive and negative events (both plotted as intensity) for a moving bee from the \emph{Bee Swarm Dataset} discussed in Section \ref{sec:experimental_setup}.
Since we assume there is no uncertainty in the polarity measurement $\sigma_k \in \{\pm 1\}$ then \eqref{eq:blob_probability} captures Gaussian uncertainty in the position of the leading and trailing edge events.
Note that with this model, should a grey object pass from a dark background to a white background then $\Delta \mapsto -\Delta$ will switch to capture the change in polarity of the visual object being tracked.

%--------------------------------%
%----- Event Blob Detection -----%
%--------------------------------%
\section{Event Blob Detection}
\label{sec:detection}

Suppose an event blob object parameterised by $\zeta(t)$ is moving across the image plane with constant translational velocity $v(t) = (v_{x}, v_{y})$.
The optical flow generated at the pixels through which the object passes should be consistent.
Objects of this form can be detected by identifying sequences of events that occur across consecutive pixels with consistent time steps between them.

The \emph{Surface of Active Events} (SAE) \cite{2013_benosman_sae} is a temporal representation for event data that is well-suited for estimating pixel-wise velocities.
It is a surface map, denoted $T$, that stores the timestamp of the most recent event to occur at a pixel.
That is, $T(x,y) = t_k$ for $e_k$ the most recent event with $\xi_k = (x,y)$.
An advantage of the SAE is that it is computationally efficient to update and maintain.
\subsection{Field of Active Flow Directions via Regression on SAE}
\label{sec:direction_estimation}
% \todo{[ZW: better to mention \emph{Field of Active Flow Directions} early in this section (even in the title).][TM: Agreed added]}
The detection criteria is based on the assumption that the object is moving in a straight line across the image, at least for a short period of time.
A key observation is that even when events occur along the line of motion, they will occur out of sequence with the progression of the object.
In particular, leading edge events occur in advance of the trailing edge events.
%rather than occurring at the same time.
%Even if only a single polarity of events are used, the spatial distribution of the image gradient of the object causes events to occur ahead or behind their expected point.
As a consequence, it is difficult or impossible to regress directly for the velocity of motion, however, it is generally quite easy to regress for a linear feature.

Imagine an object moving with constant optical flow $v$ over a short period of time.
Events will lie on, or stochastically close, to a linear feature $L(v, \xi_k)$
%\begin{subequations}\label{eq:regr_motion_equation}
%    \begin{align}
%        (t_{k} - T(x,y)) \cdot v_x &= (\xi^{x}_{k} - x)  + \mu^x_k \label{eq:regr_motion_equation_x} \\
%        (t_{k} - T(x,y)) \cdot v_y &= (\xi^{y}_{k} - y) + \mu^y_k \label{eq:regr_motion_equation_y}
%        \end{align}
%\end{subequations}
\begin{align}
    % v_\perp^\top (\xi  - \xi_k) =  0
    (\xi  - \xi_k)^\top v_\perp = 0
\label{eq:regr_motion_equation}
\end{align}
where $v_\perp \in \R^2$ with $|v_\perp| = 1$ is the orthogonal direction to the line and $\xi_k$ is a point on the line.
% \todo{[ZW: Because of the sign, the transpose of $v_\perp$ is a bit confusing. Might need to point it out.][TM: added clarification]}

Consider an event $e_k = (\xi_k, \sigma_k, t_k)$ and consider the question of whether this event is associated with a salient moving object in the image.
Choose a patch around the event location $\xi_k$ and let $\{\xi_i\}$ for $i = 1, \ldots, n$ index the pixels in the image patch, excluding $\xi_k$.
We assign a conditional probability that a pixel $\xi_{i}$ in the patch belongs to a linear feature $L(v,\xi_k)$ based on how close it is to the line, scaled by how recently an event occurred at that pixel
\[
\textbf{p}(\xi_{i} \in L(v,\xi_k) | T(\xi_{i}), v_\perp, \xi_k) =
\exp\left(
\frac{-(v_\perp^\top (\xi_{i}  - \xi_k))^2}{\exp(2\alpha\delta_k(\xi_{i}))}
\right)
\]
where $\delta_k(\xi_{i})= (t_{k} - T(\xi_{i}))$ and the rate parameter $\alpha$ is used as a tuning parameter.
Thus, recent events are modelled as stochastically close to the line, while historic events are not considered informative and are assigned very high variance.
% Set $a_i^\top = (\xi(i)^\top 1) \in \R^{3 \times 1}$ and choose $\ell_k
% = (v_\perp, v_\perp^\top \xi_k) \in \R^3$.
Set $a_i = (\xi_i  - \xi_k) \in \R^{2}$ and choose $\ell_k = v_\perp \in \R^2$.
Define the diagonal matrix
\begin{align} \label{eq:exp_W_mat}
    W =
    \diag ( e^{2\alpha  \delta_k(\xi_1)}, \ldots, e^{2\alpha  \delta_k(\xi_n)})
%    \begin{pmatrix}
%        e^{2\alpha  \delta_k(\xi(1))} & 0 & 0  \\
%%        0 & e^{2\alpha \cdot \delta_k(\xi(1))} & \hdots & 0 & 0 \\
%        0 &  \ddots & 0 \\
%%        0 & 0 & \hdots & e^{2\alpha \cdot \delta_k(\xi())} & 0 \\
%        0 & 0 & e^{2\alpha  \delta_k(\xi(n))}\\
%    \end{pmatrix}
\end{align}
and the vector $
    A_k =
    \begin{pmatrix}
        a_1  & a_2 & \hdots & a_n
    \end{pmatrix}^\top.
$
Then the parameters $\ell_k$ for the line are the solution of the least squares problem $A_k \ell_k = \mu_k$ with $\mu_k \in \GP(0,W)$.
The least squares solution to this regression is given by solving for the eigenvector $\hat{\ell}_k$ associated with the smallest eigenvalue of $(A^{\top}W^{-1}A)$.
For this, $\hat{v}_\perp = \hat{\ell}_{k}$ and the direction of the line is given by
% $
% \hat{v}_\perp = \hat{\ell}_k(1:2) | \hat{\ell}_k(1:2)|^{-1}
% $
% is the normalised two-vector from the solution and the direction of the line is given by
\[
\hat{v}_k = \pm \begin{pmatrix} 0 & -1 \\ 1 & 0 \end{pmatrix} \hat{v}_\perp \in \RP^2
\]
on the real projective space of directions.
We will choose the second element to be positive, or if that is zero, then the first element to be positive.

Define a \emph{Field of Active Flow Directions} to be an assignment $V : \R^2 \to \RP^2$ from the image to the real projective space, namely, $V(\xi_k) = \hat{v}_k$.
That is, when each event occurs, we compute the associated most likely flow direction from the Surface of Active Events and then store that in the $\xi_k$ entry of the field of active flow directions.
The field $V$ is complementary to the SAE, providing a history of the direction of optical flow estimation associated with each event at the time when it occurred, and provides a powerful cue to salient motion in the image.

%----- Classifying Detection -----%
\subsection{Detection Criteria}
\label{sec:classifying_detection}

The detection criteria used is based on correlation of the field of active flow directions with the latest flow estimate.
Using the same framework as Section \ref{sec:direction_estimation}, linear regression is used to estimate the dominant flow direction $\ob{V}$ in the flow field $V(\xi_{i})$ in the neighbourhood of event $e_{k}$. 
In the absence of noise then the flow field is parallel and there exists
$\ob{V}_\perp \in \R^2$ with $|\ob{V}_\perp| = 1$ such that 
\begin{align}
    V(\xi_i)^{\top} \ob{V}_{\perp} = 0 
\end{align}
for all $\xi_i$ in the patch. 
Like Section \ref{sec:direction_estimation}, $\ob{V}_{\perp}$ is estimated using weighted least squares regression where component $V(\xi_{i})$ is weighted by how recently it was estimated.

We propose a detection criteria $C = \lvert \langle V(\xi_{k}), \ob{V}_\perp \rangle \rvert$ as the magnitude of the inner product between the latest active flow direction $V(\xi_{k})$ at the event pixel $\xi_{k}$ and $\ob{V}_\perp$ obtained from the regression.
Small magnitude of $C = \lvert \langle V(\xi_{k}), \ob{V}_\perp \rangle \rvert$ indicates high probability of linear motion.
A candidate detection is registered if $C < \gamma$, where $\gamma$ is a threshold parameter that must be tuned.

% \vspace{5mm}
% The detection criteria used is based on correlation of the field of active flow directions with the latest flow estimate.
% For all pixels in a neighbourhood of an event $e_k$ compute
% \begin{align*}
% \ob{V}_k = \frac{1}{\sum_{i=1}^{n}\exp(-2 \alpha \delta_k(\xi_i))}
% \sum_{i=1}^{n}\exp(-2 \alpha \delta_k(\xi_i))V(\xi_i)
% \end{align*}
% % and set
% % \[
% % \ob{V}_k = \frac{\ob{V}'_k}{\|\ob{V}'_k \|}
% % \]
% that measures the average weighted direction of the field of active flow directions.
% % \cite{2009_Mardia_Direction_Stats}.
% %and the associated correlation matrix
% %\begin{align*}
% %\Sigma_k^V  = &
% %\frac{1}{\sum_{i=1}^{n}\exp(-2 \alpha \delta_k(\xi_i))} \\
% %& \sum_{i=1}^{n}\exp(-2 \alpha \delta_k(\xi_i))
% %(V(\xi_i) - \ob{V}_k)(V(\xi_i) - \ob{V}_k)^\top.
% %\end{align*}
% We propose detection criteria $C = \lVert V_{\xi_{k}} - \ob{V}_k \rVert$ as the difference between the weighted mean velocity and the latest active flow direction $V(\xi_k)$.
% A candidate detection is registered if $C < \gamma$, where $\gamma$ is a threshold parameter that must be tuned.
% Note that $C$ measures both the correspondence of $V_{\xi_k}$ with the direction $\ob{V}_k$ as well as the correlation within the active field of flow directions $V(\xi)$, since for uncorrelated fields the magnitude of $\ob{V}_k$ becomes small.

%--------------------------------%
%----- Event Blob Detection -----%
%--------------------------------%
\section{Pre-Tracking and Validation}
\label{sec:pretrack_and_validation}
When a general salient feature is detected, a candidate AEB Tracker \cite{wang2023event} is spawned with initial position $\xi_{k}$ and velocity $V(\xi_k)$.
% This is a full-function AEB Tracker, described in Section \ref{sec:tracking}, that operates in the background.
As the filter evolves, an intensity patch of the associated events, centred on the filter state, is generated.
A small neural network trained for the object of interest is used for binary classification of the intensity patch, where the classification output is used to validate the track.

\subsection{Candidate Track Intensity Patch}
\label{sec:intensity_patch_generation}

\begin{figure}[t!]
    \centering
     \begin{subfigure}[t!]{\columnwidth}
        \centering
        \includegraphics[width=\columnwidth]{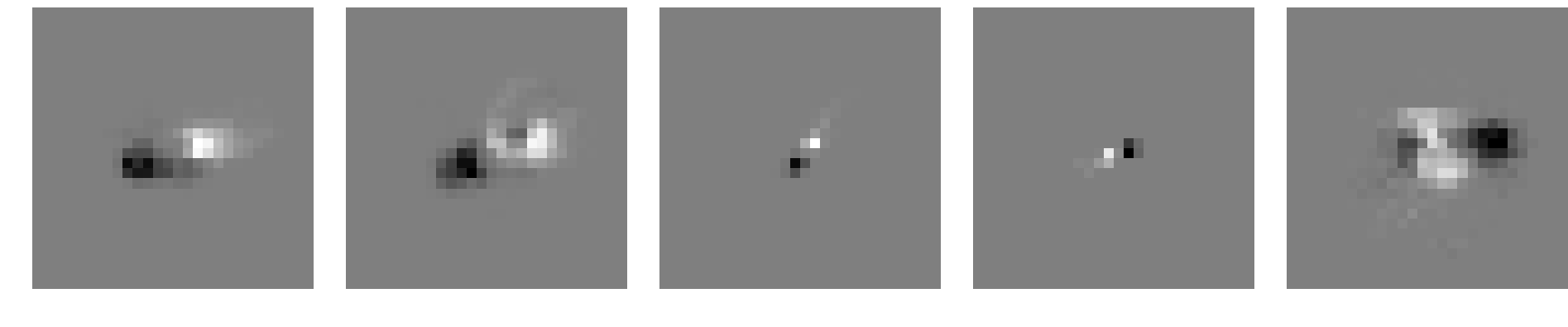}
        \vspace{-7.5mm}
        \caption{True tracks (bees)}
        \label{fig:intensity_patch_true}
    \end{subfigure}%
    \vfill
    \begin{subfigure}[t!]{\columnwidth}
        \centering
        \includegraphics[width=\columnwidth]{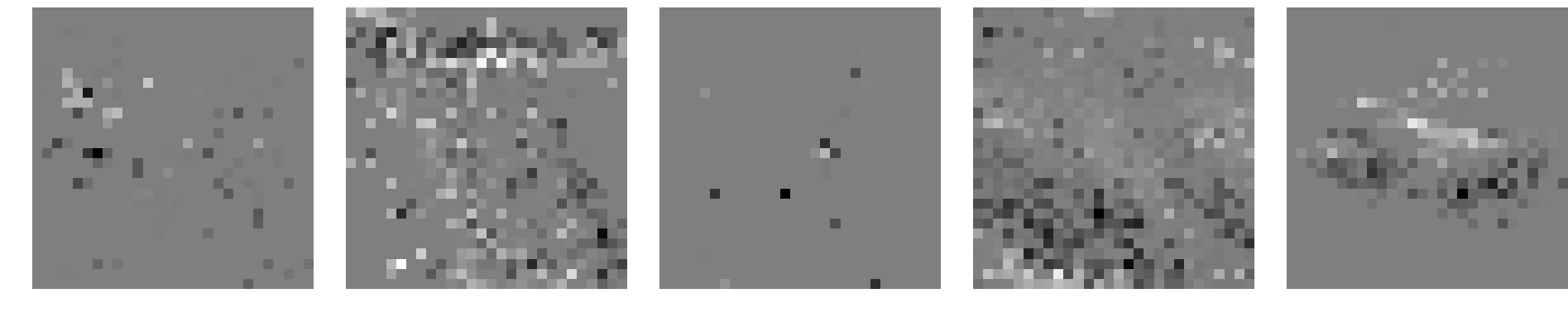}
        \vspace{-7.5mm}

        \caption{False tracks (not bees)}
        \label{fig:intensity_patch_false}
    \end{subfigure}
    \vfill
    % \vspace{-1mm}
    \caption{Example intensity patches used for track validation on the \textit{Bee Swarm Dataset}, where the intensity of events is shown. (a) shows true tracks (bees) where the leading and trailing are clear and there is a well-defined structure. (b) shows false tracks (background) where this structure is not seen and there is a more random distribution of events.}
    \label{fig:intensity_patch}
    \vspace{-6mm}
\end{figure}

For each candidate track, a 28$\times$28 pixel intensity patch $I$ is generated using the associated events to describe the structure of the object being tracked.
To align the patch with the center of the blob, each event is offset by the current position of the blob.
The lifetime of each event added to the intensity patch is managed using an exponential decay factor, which allows the structure to evolve over time by gradually forgetting past events \cite{2018_Scheerlinck_ComplementaryFilter}\cite{2023_Wang_AKF}.
For use with the \emph{Bee Swarm Dataset}, illustrative examples of true and false ``bee'' targets are shown in Figure \ref{fig:intensity_patch}.

\subsection{Intensity Patch Classification}
\label{sec:patch_classification}

Validating a track using intensity patches is posed as a binary image classification task, which is well-suited for a simple neural network.
AEMOT is designed to be general, and hence the neural network is simple so as to be easily retrained for particular objects of interest.
For application on the \textit{Bee Swarm Dataset}, we developed a fully-connected neural network for binary classification of 28$\times$28 intensity patches into ``bee" or ``not bee" categories.
The architecture includes two hidden layers: the first fully connected layer maps the 784-dimensional flattened image vector to 128 neurons, and the second layer reduces this to 64 neurons, each followed by ReLU activation functions.
Finally, a single-neuron output layer with a sigmoid activation produces a probability score for binary classification.
The network is trained using Binary Cross-Entropy Loss and the Adam optimizer on a dataset of 2,000 positive and 2,000 negative samples.
Thanks to the small dataset size and efficient architecture, the training process can be completed within 1-2 minutes on a single CPU.
The pre-trained model is then integrated with LibTorch in C++ for offline inference, ensuring a balance between accuracy and computational efficiency, making it ideal for deployment on devices with limited resources.

A small FIFO evaluation buffer of 15 entries is used to maintain a short history of the classification results.
A track is classified as a ``bee'' when all entries classify the track as ``bee''.
When this occurs, the candidate track is promoted to a valid track.
The track retains the history captured while classified as a candidate.
When all entries classify a track as non-bee, then the candidate track is terminated.

%--------------------------------%
%-----  Event Blob Tracker  -----%
%--------------------------------%
\section{Tracking and Track Management}
\label{sec:tracking}
AEB trackers \cite{wang2023event} are used and managed to track each candidate and validated object.

\subsection{Asynchronous Event Blob (AEB) Tracker}
The AEB tracker uses two pseudo measurements to estimate the state, $\zeta(t)$, of a blob, including its shape.
Pseudo-measurement abstraction is necessary since the uncertainty in the event location is related to the size of the blob, that is itself a state in the filter \cite{wang2023event}.

The first measurement function is defined as,
\begin{align*}
    H(x_{k};\xi_{k}) := \Lambda^{-1}_{k}(\xi_{k} - \sigma_{k}\Delta_{k} - p_{k})
\end{align*}
where $\Lambda$ is defined by \eqref{eq:blob_Lambda}, $\xi_{k}$ is the location of the incoming event, $\Delta$ is the polarity offset parameter, and $p_{k}$ is the event blob position.
The generative measurement model is
\begin{align}
	\label{eq:measurement_H}
0 & =  H(x_{k};\xi_{k}) + \eta_k, & \eta_k \sim \GP(0,I_2)
\end{align}
where $0$ is a pseudo-measurement and the uncertainty model can be derived from \eqref{eq:blob_probability} \cite{wang2023event}.
The second measurement is required to ensure observability of the shape parameter $\Lambda_k$,
\begin{align*}
    G(&x_{k}; \hat{P}^{-}_{k},\Xi_{k}) \\
    &:= \sum_{j=1}^{n} \lVert \frac{1}{1+\beta} \hat{\Lambda}^{-1}_{k-j}(\xi_{k-j} - \sigma_{k-j}\Delta_{k-j} - \hat{p}_{k-j}^{-})  \rVert^2
\end{align*}
where $\hat{P}^{-}_{k}$ is a buffer of $n$ past state prediction estimates $\hat{p}_{k-j}^{-}$ from the AEB tracker, $\beta$ is an upper bound on the $2$-norm of their covariances, and $\Xi_{k}$ is a buffer of associated events \cite{wang2023event}.
The generative measurement model is
\begin{align}
\label{eq:measurement_G}
2 n & =      G(x_{k};\hat{P}^{-}_{k},\Xi_{k}) + \nu_k,
& \nu_k \sim \GP(2n, 4n)
\end{align}
where $2n$ is a pseudo-measurement and the uncertainty model is a chi-squared distribution for $n$ variables.
The two-pseudo measurements \eqref{eq:measurement_H} and \eqref{eq:measurement_G} are independent \cite{wang2023event}, and the combined measurement model used is
\begin{align}
    m_{k} =&  \begin{pmatrix}
        H(x_{k}; p_{k}) \\ G(x_{k}; \hat{P}^{-}_{k}, \Xi_{k})
    \end{pmatrix} + \upsilon_k, \notag \\
    & \makebox[2cm]{} \upsilon_k \sim \GP((0,2n), \diag(I_2, 4n))
\end{align}

The standard Extended Kalman Filter (EKF) equations are used to fuse the estimated state with the incoming measurements \cite{wang2023event}, with the linearised measurement model $C_{k}$ defined by the Jacobians of the pseudo measurements,
\begin{align}
    C_{k} = \begin{pmatrix}
        C^{H}_{k} \\ C^{G}_{k}
    \end{pmatrix}
    \in \mathbb{R}^{3 \times 10}
\end{align}

%-------------------------------------------------------%
%-----               Track Management              -----%
%-------------------------------------------------------%

\begin{figure*}
    \includegraphics[width=\textwidth]{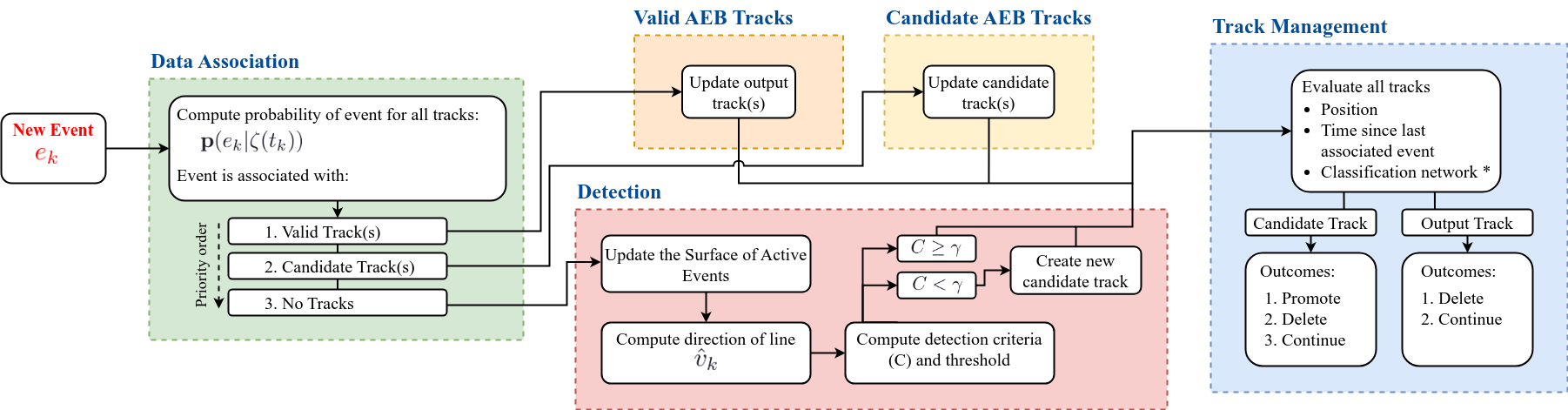}
    \vspace{-7mm}
    \caption{Block diagram of AEMOT. For each event, data association is performed (in priority order) to assign the event to any existing valid or candidate tracks, otherwise detection is performed. All tracks are evaluated using position and time since the last associated event. The classification network is evaluated periodically, which can be configured to suit the application. Following evaluation, valid tracks are deleted or continue, and candidate tracks are promoted, deleted or continue.
    % \todo{[ZW: do you want to add the bee classifier as optional?]}	
    }
    \label{fig:MTT_pipeline}
    \vspace{-7mm}
\end{figure*}

\subsection{Track Management}
%\label{sec:track_management}
The complete AEMOT algorithm is illustrated in Figure \ref{fig:MTT_pipeline} and shows how track management maintains tracks, including data association, track interaction and termination.

\subsubsection{Data Association}
\label{sec:data_association}

Data association is performed at the arrival of each new event $e_{k}$ using the event probability model \eqref{eq:blob_probability}.
For each track, the squared Mahalanobis distance is computed $d_{k}^{2} = \tilde{\xi}_k^{\top} \Lambda(t)^{-2}\tilde{\xi}_k$. 
This follows a chi-squared distribution with two degrees of freedom, $\chi_{2}^{2}$.
Distance $d_{k}^{2}$ is compared to the critical value of $\chi_{2}^{2}$ at a tuneable significance value to define how well the measured event fits the estimated blob distribution.

% Data association is performed with the arrival of each new event $e_{k}$ by evaluating the event probability model \eqref{eq:blob_probability}.
% For each track, this computes the probability of event $e_{k}$ given its current state $\zeta(t_{k})$.
% A threshold is used to determine how well an event must match the probability model of the blob to be associated.
Data association is first performed for all existing valid tracks, and if no associations are made, data association is then performed for all candidate tracks.
There are three outcomes from data association: an event can be associated with no tracks, one track, or multiple tracks.
When no tracks are associated, the event is provided to the detector.
If the event is associated to one track, the EKF prediction and update steps are performed for the track, incorporating the event as a measurement.
If the event is associated to multiple tracks, only the prediction step of the EKF is performed for all associated tracks.
By only performing prediction when the source of the event is ambiguous, tracks are forward integrated through interactions using their velocities until they are uniquely associated events again (see Figure~\ref{fig:crossing_paths} and associated discussion).

\subsubsection{Termination}
\label{sec:termination}
When an AEB tracker is no longer tracking a suitable object, it is identified quickly and terminated to avoid reporting false tracks and using computational resources.
After each event is processed, all tracks are evaluated using the position and time since the last associated events.
A track is terminated if it is no longer within the image plane or the elapsed time since the last associated event exceeds a threshold.
The elapsed time threshold is proportional to the average event rate of each track to identify discrepancies in the event rate of the track.

Tracks are evaluated periodically using the classifier described in Section \ref{sec:pretrack_and_validation}.
After a batch of 50 events are associated to a track, its intensity patch is input to the classifier and the result is stored in an evaluation buffer for that track.
If all classifications in the buffer are false, the track is terminated.

%-------------------------------------------------------%
%-----             Experimental Setup              -----%
% ------------------------------------------------------%
\section{Experimental Set-Up}
\label{sec:experimental_setup}

In this section, we detail the dataset, implementation, and performance metrics we use to examine AEMOT.

\subsection{Dataset}
\label{sec:dataset}
All experiments were performed on the \textit{Bee Swarm Dataset}, a short recording of an active bee swarm where hundreds of bees fly in all directions.
In general, bees only occupy a few pixels at a time and due to the windy weather conditions, there are very active trees in the lower section of the image that provide challenging background activity.
The data was recorded with a Prophesee EVK4 HD event camera with standard bias configuration.
The \textit{Bee Swarm Dataset} is a 5 second recording (approximately 5 million events) with an average event rate of 1Me/s.
The dataset has been manually labelled to assign each event to the corresponding bee or the background, and is subject to human error.
Caution has been taken to avoid false labelling of background as bees, resulting in some unlabelled (difficult to detect) bees.
On average, there are approximately 80 labelled bees in the image plane, ranging between 70 and 110 throughout the dataset.

\subsection{Implementation}
\label{sec:implementation}
AEMOT is implemented in C++ using \textit{Boost}, \textit{Eigen} and \textit{LibTorch} libraries.
% No multi-threading has been implemented, but there is scope to do so.
All experiments were performed on a laptop with an Intel i7-1365U 13\textsuperscript{th} Generation processor.
Source code for AEMOT will be made available.

\subsection{Performance Metrics}
\label{sec:performance_metrics}
Algorithms are evaluated by their time-average precision and recall.
Precision and recall values are computed every 5ms (200Hz) using the identified and labelled bees at that time.
% \todo{
% This frequency is chosen to align with the clustering duration of the compared algorithms.
% }
Precision is computed as the ratio of correctly identified bees to all identified bees, and recall is computed as the ratio of correctly identified bees to all labelled bees.
Precision is the most important metric as robustly rejecting false tracks is more important than detecting and tracking all the bees.

%-------------------------------------------------------%
%-----            Experimental Results             -----%
% ------------------------------------------------------%
\section{Experimental Results}
\label{sec:experimental_results}
In this section, we report the performance of AEMOT alongside two alternatives, and in several difficult cases. %Experiments presented in Section \ref{sec:experimental_results} are performed using the \textit{Bee Swarm Dataset}.
%\todo{The presented results may vary with different labelling, however the trends should persist}. % of the results should persist.

\subsection{Algorithm Comparison}

AEMOT is evaluated against Prophesee Spatter Tracker \cite{2024_Prophesee_spatter_tracker} and the jAER Rectangular Cluster Tracker \cite{2008_Delbruck_jAER}.
% These algorithms are chosen since they demonstrate strong performance for high-speed tracking of general blob-like features.
% These algorithms significantly outperform more modern algorithms on simple blob tracking data and are state-of-the-art for the problem considered. 
These algorithms are chosen since they significantly outperform more modern algorithms on simple blob tracking data and are state-of-the-art for the problem considered. 
The optical flow of the bees regularly exceeds 1000 pixels/s where the more recent algorithms ACE \cite{2018_alzugaray_ace} and HASTE \cite{2020_alzugaray_haste} have been shown to fail (see comparison study in \cite[Section VI]{wang2023event}). 
AEMOT and Prophesee are not limited by number of tracks, whilst jAER is restricted to 20 simultaneous tracks.
AEMOT is presented with and without the classification neural network.
Without the classifier, candidate tracks are evaluated using prior knowledge of the bee object to threshold position covariance, and velocity and shape state values.
% Results are presented in Table \ref{tab:algorithm_comparison}.

% From Table \ref{tab:algorithm_comparison}, AEMOT achieves the highest performance in all performance metrics.
% It records the most correct tracks in both configurations (1.5 times Prophesee \cite{2024_Prophesee_spatter_tracker} and 3 times jAER \cite{2008_Delbruck_jAER} with classifier, and 2.15 times Prophesee \cite{2024_Prophesee_spatter_tracker} and 4.5 times jAER \cite{2008_Delbruck_jAER} without the classifier) with higher precision and recall.
% AEMOT with classifier reports the highest precision, exceeding AEMOT without classifier by 19.7\%, Prophesee \cite{2024_Prophesee_spatter_tracker} by 70\% and jAER \cite{2008_Delbruck_jAER} by 57.4\%, indicating that the validation stage in AEMOT is effective at rejecting non-bee like blobs.
% For recall, AEMOT with classifier exceeds Prophesee \cite{2024_Prophesee_spatter_tracker} by 44\% and jAER \cite{2008_Delbruck_jAER} by 200\%, suggesting that initial saliency detection in AEMOT performs well in identifying areas of interest.
% As the classifier provides robustness against false tracks, there are instances where some challenging bees are rejected resulting in lower recall than AEMOT without the classifier.
% There are, however, instances where the background matches the structure of a bee and allow false candidate tracks to be validated by the classifier, leading to imperfect precision and recall.

From Table \ref{tab:algorithm_comparison}, AEMOT achieves the highest performance in all performance metrics.
It records the most correct tracks in both configurations (1.4 times Prophesee \cite{2024_Prophesee_spatter_tracker} and 3 times jAER \cite{2008_Delbruck_jAER} with classifier, and 2.66 times Prophesee \cite{2024_Prophesee_spatter_tracker} and 5.63 times jAER \cite{2008_Delbruck_jAER} without the classifier) with higher precision and recall.
AEMOT with classifier reports the highest precision, exceeding AEMOT without classifier by 20.6\%, Prophesee \cite{2024_Prophesee_spatter_tracker} by 64\% and jAER \cite{2008_Delbruck_jAER} by 51.9\%, indicating that the validation stage in AEMOT is effective at rejecting non-bee like blobs.
For recall, AEMOT with classifier exceeds Prophesee \cite{2024_Prophesee_spatter_tracker} by 37\% and jAER \cite{2008_Delbruck_jAER} by 185\%, suggesting that initial saliency detection in AEMOT performs well in identifying areas of interest.
As the classifier provides robustness against false tracks, there are instances where some challenging bees are rejected resulting in lower recall than AEMOT without the classifier.
There are, however, instances where the background matches the structure of a bee and allow false candidate tracks to be validated by the classifier, leading to imperfect precision and recall.

\begin{figure}[t!]
    \centering
     \begin{subfigure}[t!]{\columnwidth}
        \centering
        \includegraphics[width=0.8\columnwidth]{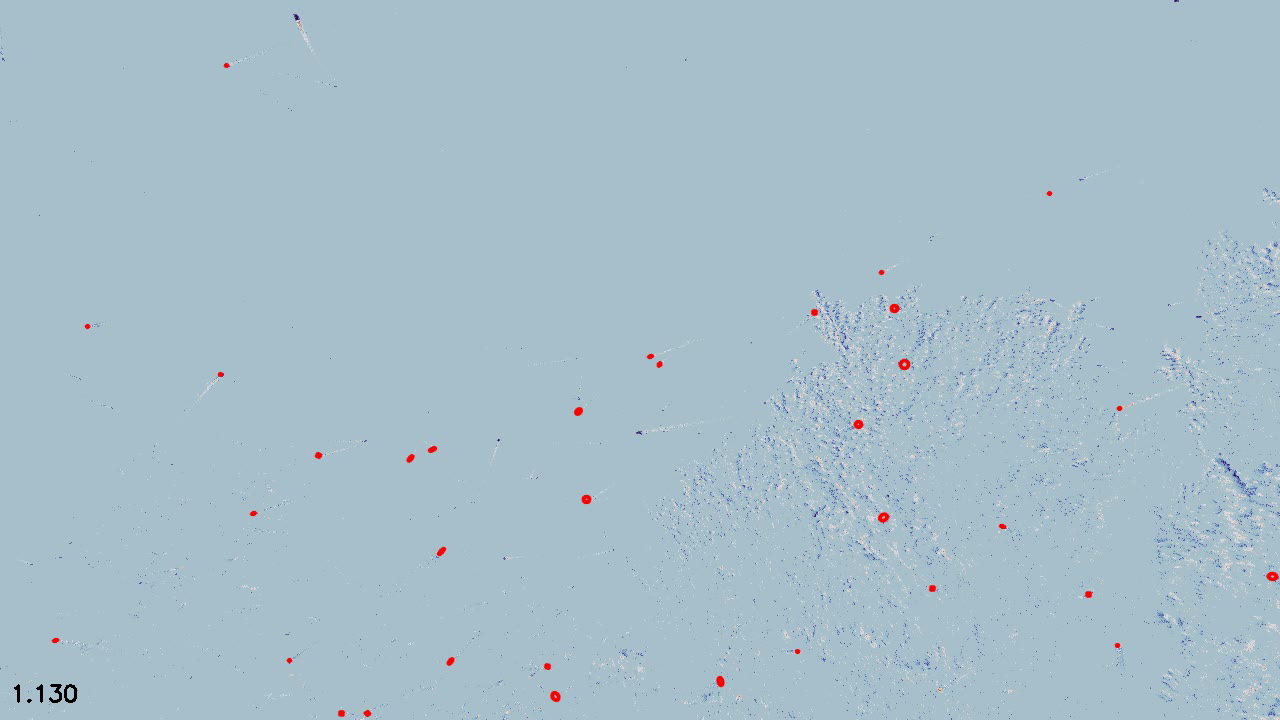}
        \vspace{-1.5mm}
        \caption{AEMOT with classifier (Ours)}
        \label{fig:aemot_sample}
    \end{subfigure}%
\vfill
     \begin{subfigure}[t!]{\columnwidth}
        \centering
        \includegraphics[width=0.8\columnwidth]{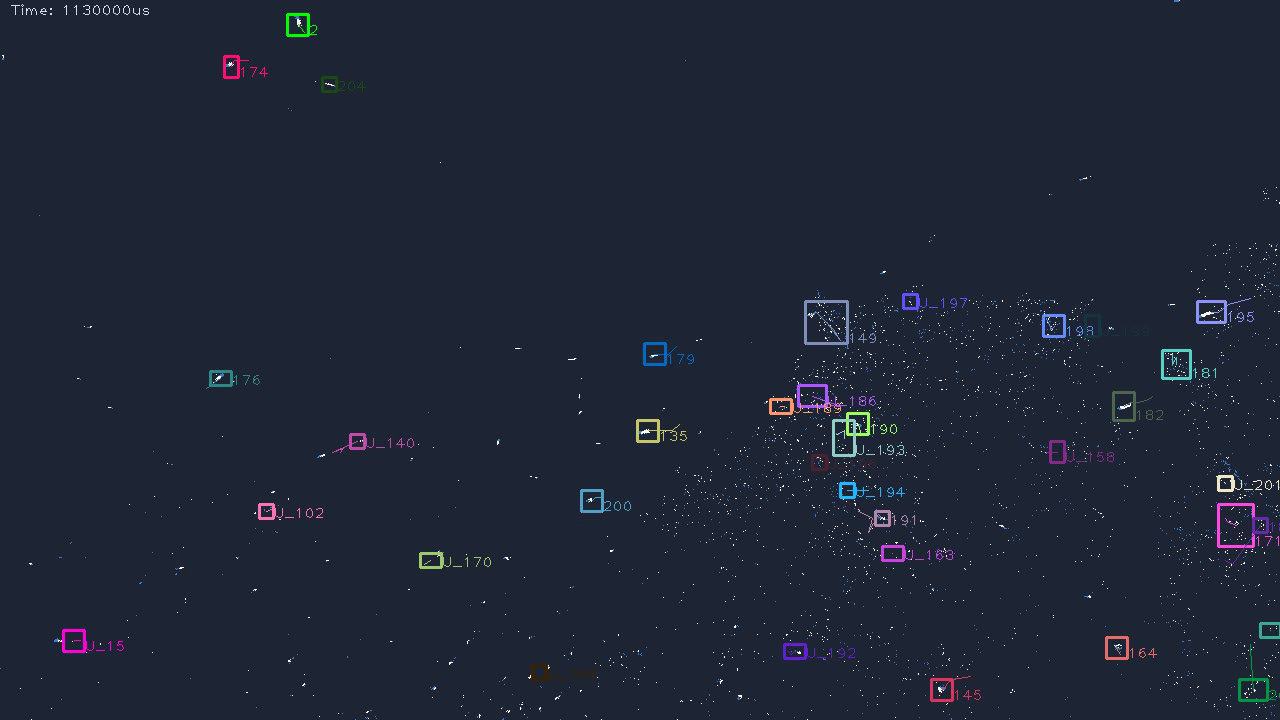}
        \vspace{-1.5mm}
        \caption{Prophesee \cite{2024_Prophesee_spatter_tracker}}
        \label{fig:prophesee_sample}
    \end{subfigure}
\vfill
     \begin{subfigure}[t!]{\columnwidth}
        \centering
        \includegraphics[width=0.8\columnwidth]{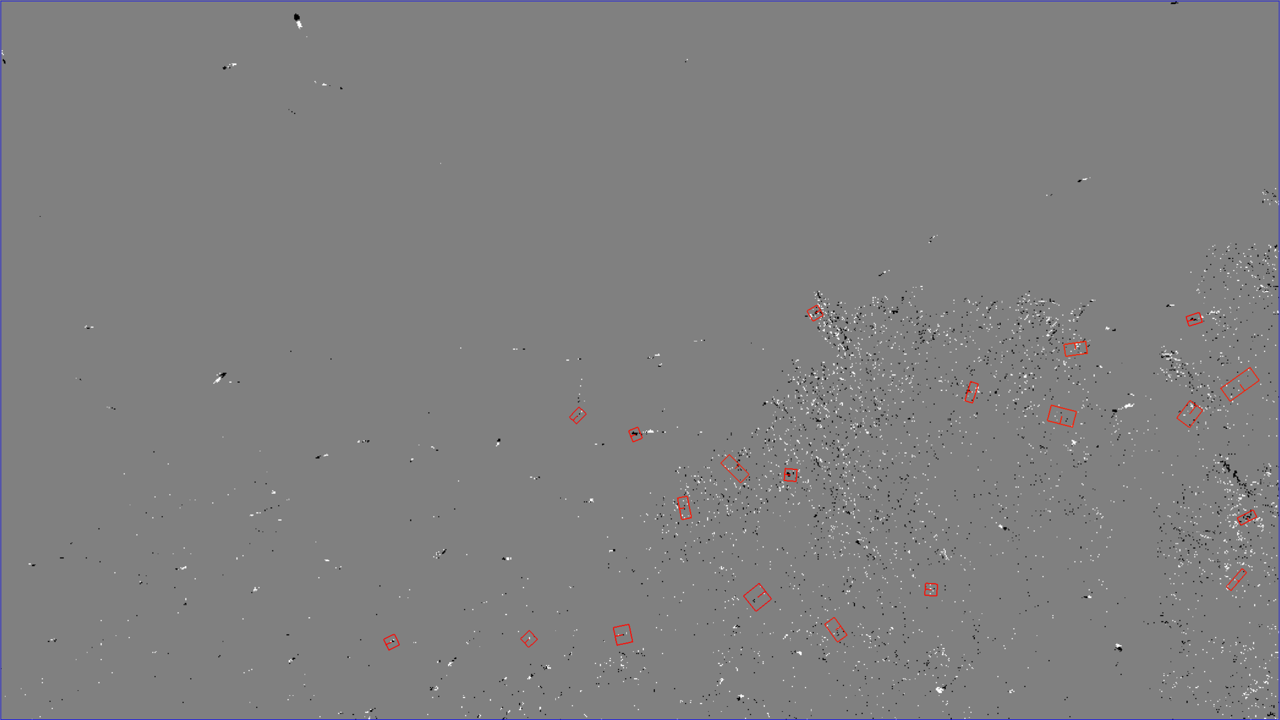}
        \vspace{-1.5mm}
        \caption{jAER \cite{2008_Delbruck_jAER}}
        \label{fig:jaer_sample}
    \end{subfigure}
    \vspace{-3.5mm}
    \caption{Sample frames (with tracks) at  $t=1.13$s for (a) AEMOT, (b) Prophesee \cite{2024_Prophesee_spatter_tracker}, and (c) jAER \cite{2008_Delbruck_jAER}. Importantly, this frame-like representation is not used in AEMOT as each event is processed asynchronously. See a detailed comparison in our supplementary video.}
    \label{fig:tracking_samples}
    % \todo{[ZW: (1) It would be good to make the bounding boxes in sub-figure (c) clear. (2) Need citations after both Prophesee and jAER method everywhere. (3) A short video for this will be very helpful!}
    \vspace{-3mm}

\end{figure}

Sample frames from each algorithm are presented in Figure \ref{fig:tracking_samples}.
The active trees in the lower-right quadrant provide challenging background activity that generates a number of false tracks from each algorithm, particularly Prophesee \cite{2024_Prophesee_spatter_tracker}.
The false tracks in this region captured by jAER \cite{2008_Delbruck_jAER} occupy a significant portion of the $20$ allowed tracks, reducing its capacity to detect and track other distinct bees (in the upper-left quadrant).
There are many small bees in the centre of the frame, which AEMOT detects and tracks better than the other algorithms.
% \todo{
% Asynchronous processing allows AEMOT to maintain robust tracks at the event rate of the received data (kilohertz).
% }
As AEMOT asynchronously models the spatial distribution of the object, it provides more accurate size and orientation estimates, where Prophesee and jAER \cite{2008_Delbruck_jAER} are limited by clustering accumulation time and bounding box representation.

\begin{table}[t!]%[H]
    \centering
    \caption{Mean (standard deviation) true and false detections, precision and recall for multi-object tracking algorithms on \textit{Bee Swarm Dataset}. Samples calculated at 200Hz. Mean (standard deviation) number of labelled tracks: 80 ($\pm$9).}
    \vspace{-2mm}
    \label{tab:algorithm_comparison}
    % \vspace{-2mm}
    \begin{tabular}{| >{\centering\arraybackslash} m{0.21\columnwidth}
                    | >{\centering\arraybackslash} m{0.125\columnwidth}
                    | >{\centering\arraybackslash} m{0.125\columnwidth}
                    | >{\centering\arraybackslash} m{0.125\columnwidth}
                    | >{\centering\arraybackslash} m{0.125\columnwidth}
                    | >{\centering\arraybackslash} m{0.125\columnwidth}|}
    \hline
    Algorithm & True Tracks & False Tracks & Precision & Recall\\
    \hline
    % \rule{0pt}{5ex} AEMOT &  \textbf{44.41} ($\pm7.65$)& \textbf{19.71} ($\pm7.45$)& \textbf{0.70} ($\pm0.08$)& \textbf{0.57} ($\pm0.11$)\\
    \rule{0pt}{5ex} AEMOT & 28.88 ($\pm7.32$)& \textbf{7.21 ($\pm4.69$)}& \textbf{0.82 ($\pm0.10$)}& 0.37 ($\pm0.11$)\\

    \hline
    \rule{0pt}{3ex} AEMOT (w/o classifier) & \textbf{55.64 ($\pm8.64$)}& 28.03 ($\pm10.77$)& 0.68 ($\pm0.08$)& \textbf{0.71 ($\pm0.11$)}\\
    \hline
    \rule{0pt}{5ex} Prophesee \cite{2024_Prophesee_spatter_tracker}& 20.93 ($\pm4.39$)& 21.87 ($\pm8.98$)& 0.50 ($\pm0.11$)& 0.27 ($\pm0.05$)\\
    \hline
    \rule{0pt}{5ex} jAER \cite{2008_Delbruck_jAER} & 9.88 ($\pm2.42$)& 8.37 ($\pm2.93$)& 0.54 ($\pm0.13$)& 0.13 ($\pm0.03$)\\
    \hline
    \end{tabular}
    \vspace{-6mm}
\end{table}

\subsection{Crossing Paths}

\begin{figure}[t!]
    \centering
     \begin{subfigure}[t!]{0.5\columnwidth}
        \centering
        \includegraphics[width=0.95\columnwidth]{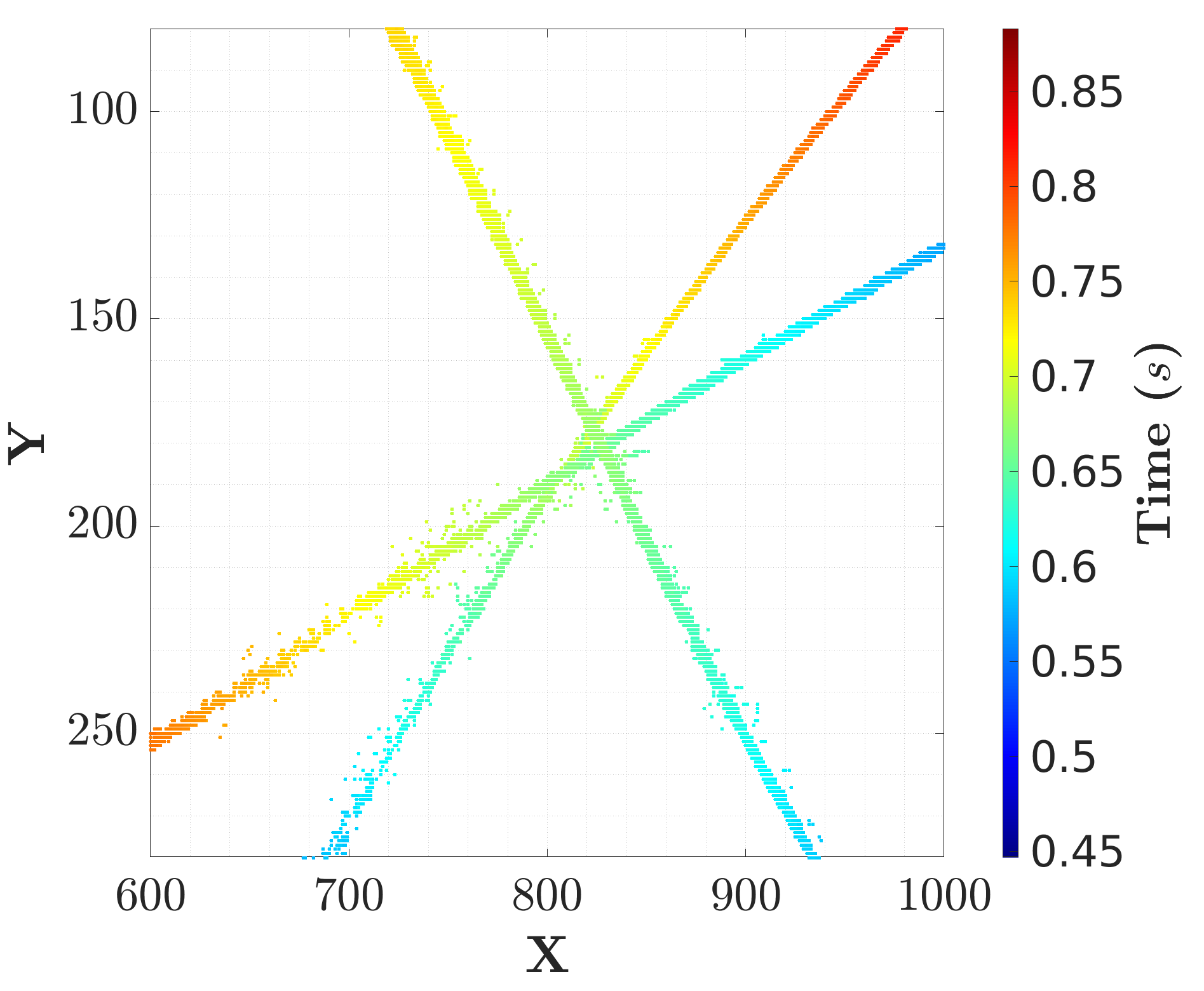}
        \vspace{-2mm}
        \caption{}
        \label{fig:crossing_paths_a}
    \end{subfigure}%
    \hfill
    \begin{subfigure}[t!]{0.5\columnwidth}
        \centering
        \includegraphics[width=0.95\columnwidth]{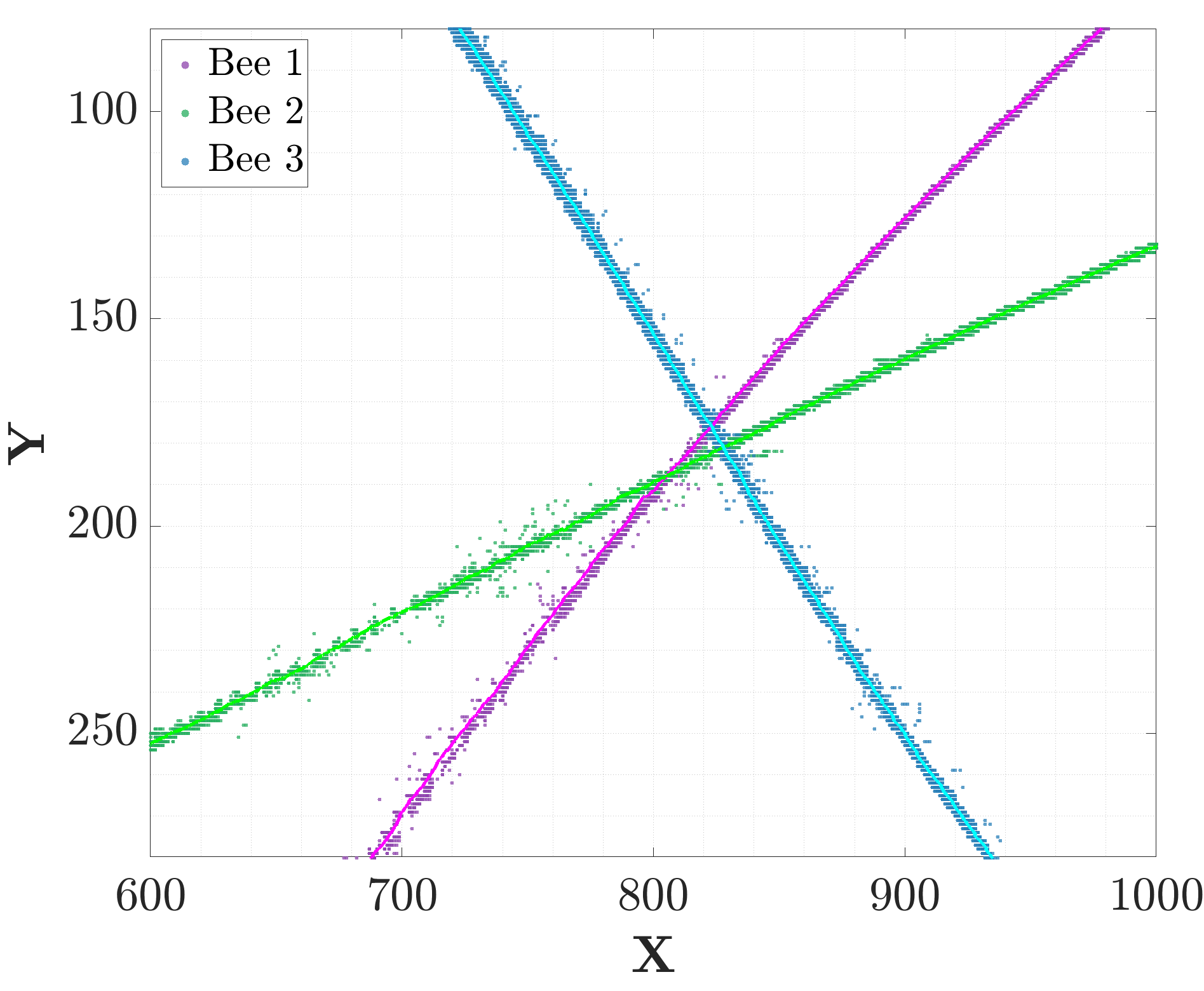}
        \vspace{-2mm}
        \caption{}
        \label{fig:crossing_paths_b}
    \end{subfigure}
\vspace{-4mm}
\caption{Example of crossing paths in \textit{Bee Swarm Dataset}, where three bees pass within close proximity. The pre-associated events (a) are shown with colour indicating the event time. The same events are shown after data association (b), with the solid lines indicating the position estimate of the respective tracks. Note that an event can be associated to multiple tracks.}
\label{fig:crossing_paths}
\vspace{-7mm}
\end{figure}

Tracking objects when their paths cross is a core challenge in multi-object tracking.
In this scenario, the source of events becomes ambiguous and discerning targets is difficult.
As detailed in Section \ref{sec:tracking}, AEMOT frames this as a data association problem identified when a single event is associated to multiple tracks, indicating they are in close proximity.
%When this occurs, only the prediction step of the EKF is performed for each track, allowing them to use their dynamic model to forward integrate through the collision until events are associated uniquely again.
Figure \ref{fig:crossing_paths} shows an illustrative example of crossing paths in the \emph{Bee Swarm Dataset}, where three bees pass very closely and two bees momentarily share the same position.
Each track is maintained through the intersection and the position estimates remain consistent.

%\section{Discussion}
%\label{sec:discussion}

%\todo{There is already a fair bit of discussion in the experimental description - I would argue we remove this section to save space
%Further data labelling is required to fully capture the bee activity in the scene and provide a more detailed understanding of the true bee objects.}

\section{Conclusion}
\label{sec:conclusion}
We introduce the novel asynchronous event multi-object tracking (AEMOT) algorithm for detecting and tracking an unknown number of objects in highly dynamic scenes.
AEMOT exploits asynchronous processing principles, the recently proposed AEB Tracker \cite{wang2023event}, and a simple neural network to achieve accurate detection and tracking at high temporal resolutions (up to the event rate of objects).
This capability is core to enabling robotic perception in unstructured real-world environments.
We demonstrate AEMOT on a newly collected \emph{Bee Swarm Dataset} that poses a significant challenge for existing event-based multi-object detectors and trackers.
Indeed, AEMOT outperforms alternative event-based detection and tracking algorithms on the bee swarm dataset by over 44\% in both precision and recall.
Both the dataset and our implementation of AEMOT will be open sourced.
Although we have trained the validation network in AEMOT for bee data, the simple neural network can be easily retrained to enable the detection and tracking of any particular objects of interest, enabling its use as an effective general feature tracker.

% An initial saliency detector identifies regions of consistent optical flow which are then evaluated using the recently proposed AEB Tracker \cite{wang2023event} and a basic neural network.
% The asynchronous model allows detection and tracking to be performed with high temporal resolution (up to the event rate of the object), which improves the capabilities of event cameras for visual perception in unstructured real-world environments.

% \todo{These conclusions are very vanilla.   The things I would like to say are something like}
% \begin{itemize}
% % \item The AEMOT algorithm exploits asynchronous processing principles to achieve high temporal resolution (up to the event rate of the object) multi-target detection and tracking of visually salient objects.
% %     This capability enables robotic perception in unstructured real-world environments.

% \item The bee data set is a challenging and powerful multi-target data set. The data set will be open source upon publication.

% \item Although we use a bee data - the algorithm is designed to be general.  The neural network is easily retrained for the particular characteristics of the blobs tracked
% \end{itemize}qq

%-------------------------------------------------------%
%-----                  References                 -----%
%-------------------------------------------------------%
\newpage
\bibliographystyle{unsrt}
\bibliography{mybib}

\end{document}

%% file: Mahony_tex_definitions/tex/preamble_compatible.tex
\usepackage{xcolor}  % Needed for colours - load before mdframed

\newcommand{\todo}[1]{{\color{red} #1}}
\usepackage{amssymb}
\usepackage{amsmath}
\usepackage{mathdots}

\usepackage{mathtools}

\usepackage{tensor}
\usepackage{xifthen}
\usepackage{urwchancal}
\DeclareFontFamily{OT1}{pzc}{}
\DeclareFontShape{OT1}{pzc}{m}{it}{<-> s * [1.10] pzcmi7t}{}
\DeclareMathAlphabet{\mathpzc}{OT1}{pzc}{m}{it}

\usepackage{graphicx}
\usepackage{ifpdf}
\ifpdf
\usepackage{epstopdf}
\PrependGraphicsExtensions{.svg}
\fi

%% file: Mahony_tex_definitions/tex/notation_defs.tex
\providecommand{\R}{\mathbb{R}}
\providecommand{\RP}{\R\mathbb{P}}

\providecommand{\GP}{\mathbf{N}} % Gaussian noise process.
\DeclareMathOperator{\diag}{diag}

\providecommand{\ob}[1]{\overline{#1}} % homogeneous vector
\usepackage{accents}
\usepackage{mathtools}
\makeatletter
\providecommand{\scirc}{%
    \hbox{\fontfamily{\rmdefault}\fontsize{0.4\dimexpr(\f@size pt)}{0}\selectfont{\raisebox{-0.52ex}[0ex][-0.52ex]{$\circ$}}}}
\makeatother
\makeatletter
\providecommand{\ucirc}{%
    \hbox{\fontfamily{\rmdefault}\fontsize{0.4\dimexpr(\f@size pt)}{0}\selectfont{\raisebox{0.0ex}[0ex][-0.52ex]{$\circ$}}}}

\makeatother

\mathchardef\mhyphen="2D
\providecommand{\idx}[5][]{
\ifthenelse{\isempty{#1}}% nothing is in the superscript spot
{\tensor*[_{#4}^{#3}]{#2}{_{#5}}}% if condition: old \idx
{\tensor*[_{#4}^{#3}]{#2}{^{#1}_{#5}}}% else condition old \ids
}

%% file: apps_detection.bbl
\begin{thebibliography}{10}

\bibitem{2011_Corke_robotics}
PI~Corke.
\newblock {\em Robotics, Vision \& Control: Fundamental Algorithms in MATLAB}.
\newblock Springer, 2011.

\bibitem{2019_falanga_fast-perception}
Davide Falanga, Suseong Kim, and Davide Scaramuzza.
\newblock How fast is too fast? the role of perception latency in high-speed
  sense and avoid.
\newblock {\em IEEE Robotics and Automation Letters}, 4(2):1884--1891, 2019.

\bibitem{2020_gallego_event-survey}
Guillermo Gallego, Tobi Delbr{\"u}ck, Garrick Orchard, Chiara Bartolozzi, Brian
  Taba, Andrea Censi, Stefan Leutenegger, Andrew~J Davison, J{\"o}rg Conradt,
  Kostas Daniilidis, et~al.
\newblock Event-based vision: A survey.
\newblock {\em IEEE transactions on pattern analysis and machine intelligence},
  44(1):154--180, 2020.

\bibitem{2021_luo_mot_review}
Wenhan Luo, Junliang Xing, Anton Milan, Xiaoqin Zhang, Wei Liu, and Tae-Kyun
  Kim.
\newblock Multiple object tracking: A literature review.
\newblock {\em Artificial intelligence}, 293:103448, 2021.

\bibitem{2011_drazen_particles}
David Drazen, Patrick Lichtsteiner, Philipp H{\"a}fliger, Tobi Delbr{\"u}ck,
  and Atle Jensen.
\newblock Toward real-time particle tracking using an event-based dynamic
  vision sensor.
\newblock {\em Experiments in Fluids}, 51:1465--1469, 2011.

\bibitem{2020_wang_particle-velocimetry}
Yuanhao Wang, Ramzi Idoughi, and Wolfgang Heidrich.
\newblock Stereo event-based particle tracking velocimetry for 3d fluid flow
  reconstruction.
\newblock In {\em Computer Vision--ECCV 2020: 16th European Conference,
  Glasgow, UK, August 23--28, 2020, Proceedings, Part XXIX 16}, pages 36--53.
  Springer, 2020.

\bibitem{2020_falanga_avoidance}
Davide Falanga, Kevin Kleber, and Davide Scaramuzza.
\newblock Dynamic obstacle avoidance for quadrotors with event cameras.
\newblock {\em Science Robotics}, 5(40):eaaz9712, 2020.

\bibitem{2020_sanket_evdodgenet}
Nitin~J Sanket, Chethan~M Parameshwara, Chahat~Deep Singh, Ashwin~V
  Kuruttukulam, Cornelia Ferm{\"u}ller, Davide Scaramuzza, and Yiannis
  Aloimonos.
\newblock Evdodgenet: Deep dynamic obstacle dodging with event cameras.
\newblock In {\em 2020 IEEE International Conference on Robotics and Automation
  (ICRA)}, pages 10651--10657. IEEE, 2020.

\bibitem{2013_delbruck_robotic-goalie}
Tobi Delbruck and Manuel Lang.
\newblock Robotic goalie with 3 ms reaction time at 4\% cpu load using
  event-based dynamic vision sensor.
\newblock {\em Frontiers in neuroscience}, 7:223, 2013.

\bibitem{2016_kim_tracking-and-reconstruction}
Hanme Kim, Stefan Leutenegger, and Andrew~J Davison.
\newblock Real-time 3d reconstruction and 6-dof tracking with an event camera.
\newblock In {\em Computer Vision--ECCV 2016: 14th European Conference,
  Amsterdam, The Netherlands, October 11-14, 2016, Proceedings, Part VI 14},
  pages 349--364. Springer, 2016.

\bibitem{2022_mahlknecht_vio}
Florian Mahlknecht, Daniel Gehrig, Jeremy Nash, Friedrich~M Rockenbauer,
  Benjamin Morrell, Jeff Delaune, and Davide Scaramuzza.
\newblock Exploring event camera-based odometry for planetary robots.
\newblock {\em IEEE Robotics and Automation Letters}, 7(4):8651--8658, 2022.

\bibitem{2022_hu_ecdt}
Sumin Hu, Yeeun Kim, Hyungtae Lim, Alex~Junho Lee, and Hyun Myung.
\newblock ecdt: Event clustering for simultaneous feature detection and
  tracking.
\newblock In {\em 2022 IEEE/RSJ International Conference on Intelligent Robots
  and Systems (IROS)}, pages 3808--3815. IEEE, 2022.

\bibitem{2018_alzugaray_ace}
Ignacio Alzugaray and Margarita Chli.
\newblock Ace: An efficient asynchronous corner tracker for event cameras.
\newblock In {\em 2018 International Conference on 3D Vision (3DV)}, pages
  653--661. IEEE, 2018.

\bibitem{2020_alzugaray_haste}
Ignacio Alzugaray and Margarita Chli.
\newblock Haste: multi-hypothesis asynchronous speeded-up tracking of events.
\newblock In {\em 31st British Machine Vision Virtual Conference (BMVC 2020)},
  page 744. ETH Zurich, Institute of Robotics and Intelligent Systems, 2020.

\bibitem{wang2023event}
Ziwei Wang, Timothy Molloy, Pieter van Goor, and Robert Mahony.
\newblock Asynchronous blob tracker for event cameras.
\newblock {\em IEEE Transactions on Robotics}, 2024.

\bibitem{2010_schraml_clustering}
Stephan Schraml and Ahmed~Nabil Belbachir.
\newblock A spatio-temporal clustering method using real-time motion analysis
  on event-based 3d vision.
\newblock In {\em 2010 IEEE Computer Society Conference on Computer Vision and
  Pattern Recognition-Workshops}, pages 57--63. IEEE, 2010.

\bibitem{2018_barranco_clustering}
Francisco Barranco, Cornelia Fermuller, and Eduardo Ros.
\newblock Real-time clustering and multi-target tracking using event-based
  sensors.
\newblock In {\em 2018 IEEE/RSJ International Conference on Intelligent Robots
  and Systems (IROS)}, pages 5764--5769. IEEE, 2018.

\bibitem{2016_brandli_elised}
Christian Br{\"a}ndli, Jonas Strubel, Susanne Keller, Davide Scaramuzza, and
  Tobi Delbruck.
\newblock Elised—an event-based line segment detector.
\newblock In {\em 2016 Second International Conference on Event-based Control,
  Communication, and Signal Processing (EBCCSP)}, pages 1--7. IEEE, 2016.

\bibitem{2016_vasco_fast-Harris}
Valentina Vasco, Arren Glover, and Chiara Bartolozzi.
\newblock Fast event-based harris corner detection exploiting the advantages of
  event-driven cameras.
\newblock In {\em 2016 IEEE/RSJ international conference on intelligent robots
  and systems (IROS)}, pages 4144--4149. IEEE, 2016.

\bibitem{2017_mueggler_fast-corners}
Elias Mueggler, Chiara Bartolozzi, and Davide Scaramuzza.
\newblock Fast event-based corner detection.
\newblock 2017.

\bibitem{2016_glover_hough}
Arren Glover and Chiara Bartolozzi.
\newblock Event-driven ball detection and gaze fixation in clutter.
\newblock In {\em 2016 IEEE/RSJ International Conference on Intelligent Robots
  and Systems (IROS)}, pages 2203--2208. IEEE, 2016.

\bibitem{2019_manderscheid_learnt-corners}
Jacques Manderscheid, Amos Sironi, Nicolas Bourdis, Davide Migliore, and
  Vincent Lepetit.
\newblock Speed invariant time surface for learning to detect corner points
  with event-based cameras.
\newblock In {\em Proceedings of the IEEE/CVF Conference on Computer Vision and
  Pattern Recognition}, pages 10245--10254, 2019.

\bibitem{2023_gehrig_recurrent-vision-transformers}
Mathias Gehrig and Davide Scaramuzza.
\newblock Recurrent vision transformers for object detection with event
  cameras.
\newblock In {\em Proceedings of the IEEE/CVF conference on computer vision and
  pattern recognition}, pages 13884--13893, 2023.

\bibitem{2023_lui_detection}
Bingde Liu, Chang Xu, Wen Yang, Huai Yu, and Lei Yu.
\newblock Motion robust high-speed light-weighted object detection with event
  camera.
\newblock {\em IEEE Transactions on Instrumentation and Measurement}, 72:1--13,
  2023.

\bibitem{2013_benosman_sae}
Ryad Benosman, Charles Clercq, Xavier Lagorce, Sio-Hoi Ieng, and Chiara
  Bartolozzi.
\newblock Event-based visual flow.
\newblock {\em IEEE transactions on neural networks and learning systems},
  25(2):407--417, 2013.

\bibitem{2018_Scheerlinck_ComplementaryFilter}
Cedric Scheerlinck, Nick Barnes, and Robert Mahony.
\newblock Continuous-time intensity estimation using event cameras.
\newblock In {\em Asian Conference on Computer Vision}, pages 308--324.
  Springer, 2018.

\bibitem{2023_Wang_AKF}
Ziwei Wang, Yonhon Ng, Cedric Scheerlinck, and Robert Mahony.
\newblock An asynchronous linear filter architecture for hybrid event-frame
  cameras.
\newblock {\em IEEE Transactions on Pattern Analysis and Machine Intelligence},
  2023.

\bibitem{2024_Prophesee_spatter_tracker}
Metavision: Applications and tools.
\newblock Accessed on September 7, 2024.

\bibitem{2008_Delbruck_jAER}
T~Delbruck.
\newblock Frame-free dynamic digital vision.
\newblock In {\em International Symposium on Secure-Life Electronics},
  volume~1, pages 21--26. University of Tokyo, March 2008.
\newblock In: Proceedings of International Symposium on Secure-Life
  Electronics, Advanced Electronics for Quality Life and Society, Univ. of
  Tokyo, Mar. 6-7, 2008.

\end{thebibliography}
